\documentclass[10pt,twocolumn,letterpaper]{article}

\usepackage{iccv}
\usepackage{times}
\usepackage{graphicx}
\usepackage{amsmath}
\usepackage{amssymb}

\usepackage{arydshln}

\usepackage{wrapfig}
\usepackage{algorithm}
\usepackage{algpseudocode}
\usepackage{soul}
\usepackage{color}
\usepackage{circledsteps}
\usepackage{pifont}
\usepackage{multirow}
\usepackage{amsfonts}       
\usepackage{nicefrac}       
\usepackage{microtype}      
\usepackage{xcolor}   
\usepackage{enumitem}
\usepackage{xspace}
\usepackage{caption}
\usepackage{subcaption}

\usepackage[accsupp]{axessibility}

\usepackage[pagebackref=true,breaklinks=true, colorlinks,bookmarks=false]{hyperref}

\definecolor{darkgreen}{rgb}{0.09, 0.45, 0.27}
\definecolor{alizarin}{rgb}{0.82, 0.1, 0.26}
\definecolor{new_cyan}{rgb}{0.10, 0.62, 0.57}

\newcommand{\QA}{Q\&A\xspace}

\usepackage[capitalize]{cleveref}
\crefname{section}{Sec.}{Secs.}
\Crefname{section}{Section}{Sections}
\Crefname{table}{Table}{Tables}
\crefname{table}{Tab.}{Tabs.}

\iccvfinalcopy 

\def\iccvPaperID{9696} 

\ificcvfinal\fi

\begin{document}
\title{Simple Baselines for Interactive Video Retrieval with Questions and Answers}

\author{Kaiqu Liang\\
Princeton University\\
{\tt\small kl2471@princeton.edu}
\and
Samuel Albanie\\
University of Cambridge\\
{\tt\small sma71@cam.ac.uk}
}

\maketitle
\ificcvfinal\thispagestyle{empty}\fi

\begin{abstract}
To date, the majority of video retrieval systems have been optimized for a ``single-shot'' scenario in which the user submits a query in isolation, ignoring previous interactions with the system.
Recently, there has been renewed interest in interactive systems to enhance retrieval, but existing approaches are complex and deliver limited gains in performance. In this work, we revisit this topic and propose several simple yet effective baselines for interactive video retrieval via question-answering. 
We employ a VideoQA model to simulate user interactions and show that this enables the productive study of the interactive retrieval task without access to ground truth dialogue data.
Experiments on MSR-VTT, MSVD, and AVSD show that our framework using question-based interaction significantly improves the performance of text-based video retrieval systems. 
Code is available at \href{https://github.com/kevinliang888/IVR-QA-baselines}{https://github.com/kevinliang888/IVR-QA-baselines}.

\end{abstract}


\section{Introduction}
\label{sec:intro}
Given the surging popularity of video content, there is a pressing need to develop tools that enable users to search video collections accurately and efficiently.
The \textit{text-video retrieval} task aims to meet this need by ranking videos among a gallery according to how well they match a given text query.
This task has seen tremendous progress in recent years, benefiting from large-scale pretraining datasets~\cite{miech2019howto100m,luo2022clip4clip}, multimodal architectures~\cite{miech2018learning,liu2019use, gabeur2020multi} and robust optimisation strategies for cross-modal representation learning~\cite{miech2020end}.
\begin{figure}[ht]
  \centering
   \includegraphics[width=\linewidth]{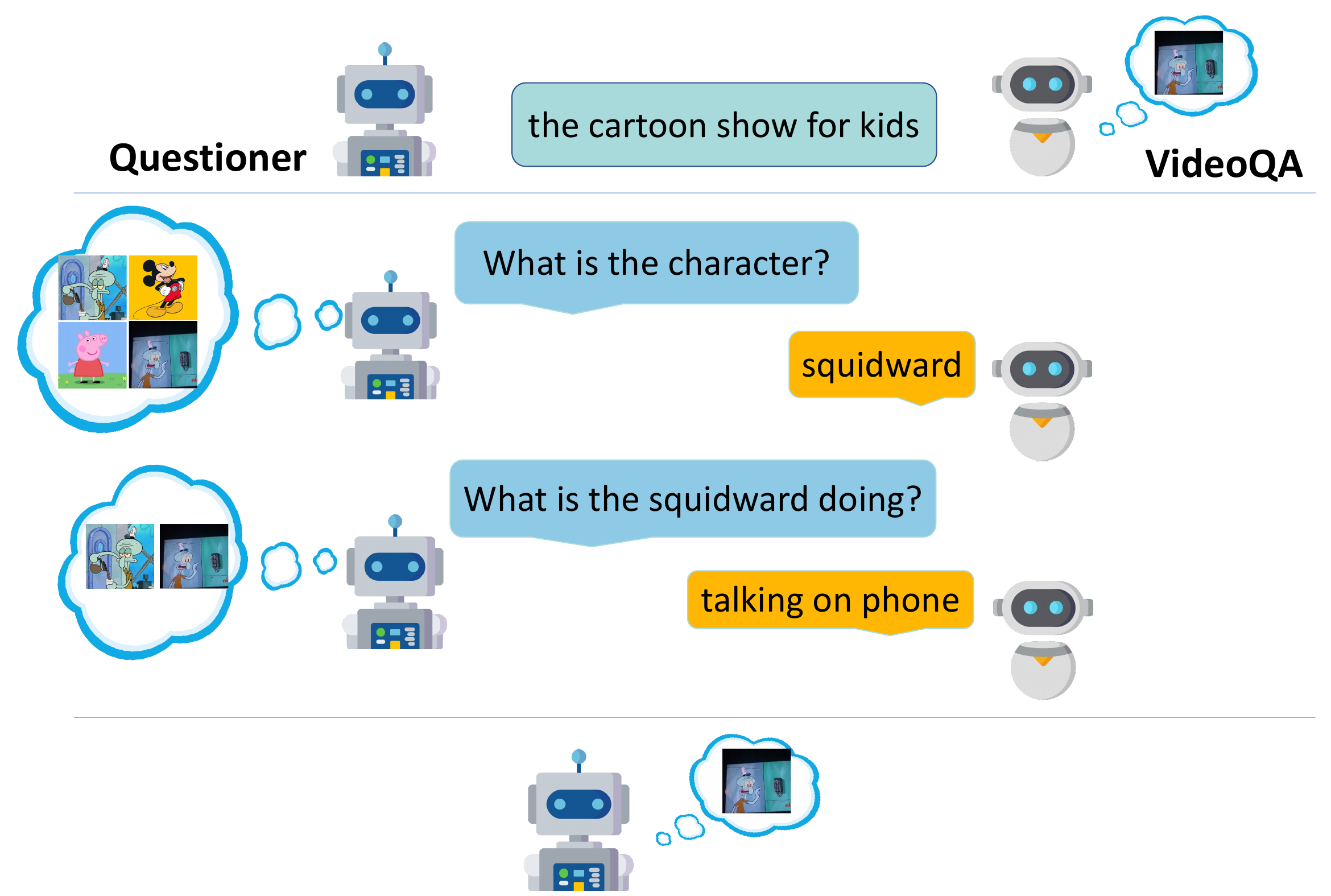}
   \caption{
   \textbf{Interactive video retrieval with questions and answers.}
   We revisit interactive video retrieval with a simple framework that employs Video Question Answering (VideoQA) to simulate user responses, sidestepping the need to collect dialogue data for development and evaluation. 
   }
   \label{fig:teaser}
\end{figure}

Much of the work to date has focused on a ``single-shot'' scenario in which a user submits a single query in isolation.
While this may be applicable in some scenarios, there are many in which it would be useful to allow refinements to the initial ranking (particularly when the pool of valid responses to the initial query is large).
To a large degree, this focus single-shot systems can be explained by practicality: obtaining the data required to develop and evaluate interactive retrieval systems is extremely challenging.
User studies that engage individuals in interactive sessions are difficult to scale, while production data from product deployments is challenging to adapt to the development of new systems.

Although it has received less attention, the topic of interactive retrieval is far from new---it has been studied for more than two decades~\cite{rui1998relevance}.
Following an initial query, various mechanisms have been offered to the user to enable them to refine results:
relevance scores~\cite{rui1998relevance} (corresponding to highly relevant, irrelevant, etc.), relative attributes~\cite{kovashka2012whittlesearch} (e.g. ``shiner than these''), natural language descriptions~\cite{guo2018dialog,tan2019drill} (e.g. ``unlike the given result, the one I want has fur on the back'')
and question-answer dialog~\cite{das2017learning,madasu2022learning} (e.g. ``Q: are there any people in the shot?'' A: No'').

Among dialogue-based approaches, the recent work of Madasu et al.~\cite{madasu2022learning} highlighted the potential of question-answering with free-form text questions to enhance video retrieval performance.
At the heart of this approach is a question-generating system that probes the user for more details about their desired target in order to refine the retrieval ranking.
In such a framework, user responses can be simulated by employing a question-answering model to synthesise plausible answers to the questions (\cref{fig:teaser}).

Our work draws inspiration from ViReD~\cite{madasu2022learning}, but differs in two key aspects.
(1) The ViReD answer generator works by first captioning, then extracting the answer from the text of the caption. The disadvantage of this approach is that the answer indirectly depends on the visual content. Instead, we adopt a video question-answering model to simulate the answer, incorporating the information from the video directly with the question.
(2) ViReD fine-tunes on AVSD with the human-collected dialogue while our approach does not require any fine-tuning. We show that our approach generalises well to multiple datasets, and obtains significantly better performance than ViReD. 

In this work, we make the following contributions:
(1) We propose a simple, yet effective framework for interactive video retrieval based on video question answering, and explore the design space of question generators within this framework.
(2) We develop three simple question generators that leverage heuristics and learned models to gather discriminative information beyond objects (scenes, actions, colour etc.) in the target video.
(3) Through careful experiments, we demonstrate that our system generalizes well to several video datasets and obtains significant improvement over baseline without fine-tuning.
These results, which can be interpreted as leveraging an approximate interactive oracle, highlight the potential of question-based interactive retrieval frameworks to substantially enhance video search. 
\section{Related work}

\noindent
\textbf{Text-to-video Retrieval} 
A long-running theme of text-video retrieval research has been the development of effective fusion mechanisms for cross-modal learning~\cite{yu2017end, kaufman2017temporal, kiros2014unifying, yu2018joint, torabi2016learning}.
Another theme of vision-language retrieval has sought to tackle challenges associated with hubness~\cite{radovanovic2010hubs} in the embedding space~\cite{dinu2014improving,bogolin2022cross}.
In the past few years, there has been increasing emphasis on the use of pre-trained models that enable retrieval systems to leverage large quantities of training data that is available for related tasks, often spanning multiple modalities~\cite{liu2019use, miech2018learning, gabeur2020multi,croitoru2021teachtext}.
Recently, there has been a particular focus on building atop models that can leverage the most scalable forms of pretraining data (e.g. CLIP \cite{radford2021learning}), which have demonstrated compelling performance across a range of vision and language tasks.
Luo \text {\it{et al.}}~\cite{luo2021clip4clip} proposed CLIP4CLIP which adapts CLIP for text-video retrieval through a combination of architectural changes and finetuning.
Other approaches have included the use of a curriculum to learn from images and video~\cite{bain2021frozen}, BERT-style architectures~\cite{sun2019videobert}, and additional pretraining tasks~\cite{li2022align}. Our approach is built upon BLIP \cite{Li2022BLIPBL}, a unified framework for retrieval, question answering, and captioning. 
Differently from the works listed above, we focus on the interactive retrieval setting.

\noindent
\textbf{Visual Question Answering. }
\noindent The goal of visual question-answering systems is to answer questions posed in natural language that concern visual content in the form of an image or video. A broad range of work in this space has sought to combine spatial image representations and sequential question representations~\cite{anderson2018bottom, ben2017mutan, fukui2016multimodal, lu2016hierarchical, xiong2016dynamic, xu2016ask, yang2016stacked}. 
To answer questions about video content, spatio-temporal video representations that encode motion and appearance have been studied in detail~\cite{fan2019heterogeneous, gao2018motion, huang2020location, jang2017tgif, jiang2020divide, jiang2020reasoning, le2020hierarchical, lei2021less, zha2019spatiotemporal}. 
There has also been work that extends these ideas to open-ended visual question answering \cite{hu2018learning, yang2021just}. 
Recent work on goal-oriented visual dialog \cite{das2017visual, das2017learning, lee2018answerer, cogswell2020dialog, padmakumar2021dialog} uses policy-based reinforcement learning \cite{padmakumar2021dialog, das2017visual, das2017learning} or optimizing information gain \cite{lee2018answerer} to improve the performance.
However, these works require intensive training on human-annotated dialog datasets and do not trivially generalise to video. 
In terms of VideoQA, a few recent works have explored zero-shot approaches \cite{yang2021just, zellers2022merlot}, where models are only trained on automatically mined video clips with short text descriptions. 
Our VideoQA model is based on BLIP \cite{Li2022BLIPBL} which is trained on manually annotated image-question-answer triplets.

\noindent
\textbf{Interactive cross-modal retrieval} 
A number of interactive image retrieval systems have been developed, drawing inspiration from visual dialogue~\cite{wu2004willhunter, rui1998relevance, kovashka2012whittlesearch, kovashka2017attributes, kovashka2013attribute, parikh2011relative, yang2020glance}. 
In such systems, users provide feedback to an agent to guide it toward the desired retrieval target. 
Two families of feedback have been studied in detail: \textit{relevance feedback} and \textit{difference feedback}.
For the former~\cite{wu2004willhunter, rui1998relevance}, users provide a relevance score for the current retrieval results. 
The system then re-ranks its initial results using the feedback provided by the user.
For the latter approach~\cite{kovashka2012whittlesearch, kovashka2017attributes, kovashka2013attribute, parikh2011relative, yang2020glance}, users communicate the difference between the target image and a reference image to the system with cues such as tags or descriptions. 
The system then ``whittles away'' irrelevant images, iteratively filtering to select the desired target. 
However, one drawback of these systems is that they place a heavy time cost on the user who must describe what is missing in the initial search results. 
For example, description-based methods~\cite{guo2018dialog, tan2019drill} require users to provide detailed sentence-level feedback and tag-based methods require users to provide a collection of attributes \cite{kovashka2013attribute, kovashka2017attributes, kovashka2012whittlesearch}.

To address these concerns, Cai \text{\it{et al}}~\cite{cai2021ask} propose a partial query system in which potentially discriminative objects are suggested to the user with the goal of narrowing down the search results quickly. 
However, this system doesn't utilize other cues in the image such as object color, actions, scenes, background, etc., so it can require many interactions with the user to obtain good retrieval results. 
By using free-form questions to elicit information from the user, our system avoids this restriction to solely object-based queries. 

Most related to our approach, the ViReD work of Madasu et al.~\cite{madasu2022learning} highlighted the potential of question-answering with free-form text questions to enhance video retrieval performance. 
However, rather than performing VideoQA directly, ViReD employs captioning followed by text-based question-answering to answer questions about the video, bottlenecking answer quality. In addition, the ViReD question generator is complex and dependent on dialog history for training---our approach requires no fine-tuning and generalises effectively to multiple datasets.
\section{Method}

\subsection{Interactive video retrieval with \QA}

\begin{figure*}[ht]
  \centering
   \includegraphics[width=\linewidth]{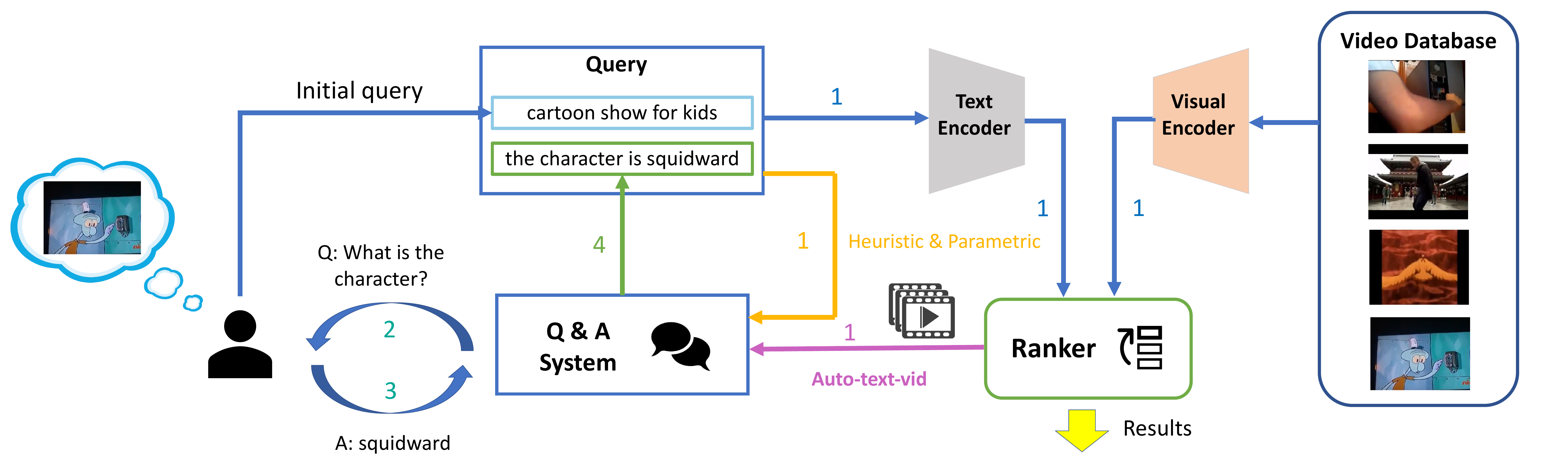}
    \caption{\textbf{Model architecture of interactive video retrieval with \QA (Questions and Answers)}. 
    The user provides an initial textual query to the system. The interaction loop contains 4 main steps.
    Step 1: Generate input to the \QA system. 
    For Heuristic and Auto-text question generation approaches, the input to the \QA system are just the queries. 
    For Auto-text-vid question generation approach, the input contains both queries and the candidate videos. 
    Step 2: Question generation.
    Step 3: User answer simulation by VQA model. 
    Step 4: New query generation. This fulfills one round of the interaction.
    }
   \label{fig:workflow}
\end{figure*}

\noindent\textbf{Components.}
Our interactive video retrieval system contains four components: a \textit{Text Encoder}, a \textit{Visual Encoder}, a \textit{Ranker}, and a \textit{\QA System}.
The Text Encoder and Visual Encoder embed partial queries and videos into a shared textual-visual space.  
The Ranker computes the similarity scores between text and video features which are used to rank the videos. 
To accelerate inference, we adopt the approach of~\cite{li2021align} that first selects $K$ candidates based on the video-text feature similarity, and then reranks the selected candidates based on their pairwise Image-Text Matching (ITM) scores. Specifically, the model uses an ITM head (a linear layer) to predict the score indicating whether an image-text pair is positive (matched) or negative (unmatched) given their multimodal feature.

\begin{figure}[ht]
  \centering
   \includegraphics[width=\linewidth]{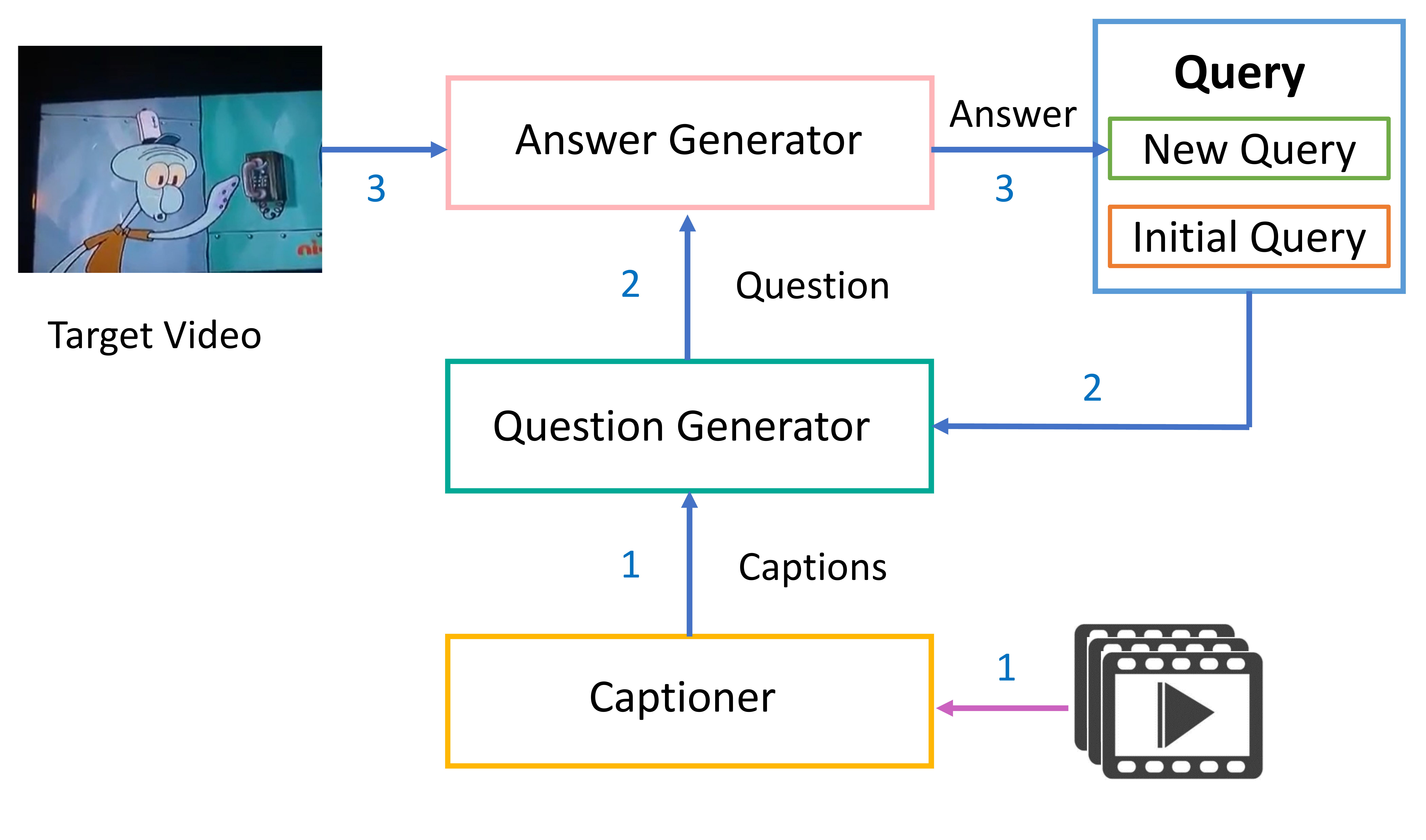}
   \caption{
   \textbf{Model architecture of \QA System}. 
   The interaction loop contains either two or three steps depending on which question generation approach is used.
   Step 1: Caption generation. 
   Step 2: Question generation. 
   Step 3: Answer generation. 
   Auto-text-vid approach contains step 1, 2 and 3 while Heuristic and Auto-text question generation approaches contain only step 2 and 3. 
   }
   \label{fig:qas}
\end{figure}

The \QA System forms the central module in the framework (\cref{fig:qas}).
It comprises a captioner adopted from the BLIP caption model, a question generator, and an answer generator that takes the form of a VideoQA (video question-answering) model that simulates user responses.
We adopt the BLIP VideoQA model~\cite{Li2022BLIPBL}.
Instead of formulating VideoQA as a multi-answer classification task \cite{Chen2020UNITERUI, Li2020OscarOA}, Li \text{\it{et al}} \cite{Li2022BLIPBL} interpret it as an answer generation task, enabling open-ended VideoQA. 
This open-ended VideoQA allows more freedom to generate and combine answers, lending flexibility to the system.

\noindent\textbf{Interaction workflow. }
The complete interaction process is demonstrated in \cref{fig:workflow}.
The user provides an initial text query to the system. 
The interaction loop then begins, comprising four main steps.
(1) \textit{Generate input to the \QA System}: 
The current (partial) query together with candidate retrieved videos is passed to the \QA system.
To obtain the candidate videos, we first generate the text embedding corresponding to the partial query and video embeddings for all videos.
The Ranker then computes similarity scores in embedding space (via cosine similarity) to rank the videos.
We select as candidate videos those who fall within the top $K$ places of the ranking and feed them as input to the \QA System. 
(2) \textit{Question generation}: We explore several kinds of question generators (described in detail in~\cref{subsec:heuristic-question-generator} and~\cref{subsec:auto-question-generator}).
Different question generators require access to different inputs.
For example, for the \textit{Auto-text-vid} question generation approach (\cref{subsec:auto-question-generator}),
we first generate captions for the candidate videos, then we generate questions based on both the captions and the partial query.
(3) \textit{User answer simulation by VideoQA model}: Given the question and the target videos, we use the VideoQA model to generate an answer. 
(4) \textit{New query generation}:  We generate a complete sentence by incorporating the answer and adding it to the original textual query. 
This completes one round of interaction.

\noindent\textbf{Combining responses after each round of interaction}
After each round of interaction, we obtain new information with the potential to improve retrieval performance.
It is therefore worth considering how this information is best integrated.
One approach is to treat each new piece of information as a separate query, then adopt a strategy for multi-query retrieval.
In recent work, Wang \text {\it{et al}}~\cite{wang2022multi} propose two \textit{post hoc} inference methods, \textit{Similarity aggregation (SA)} and \textit{Rank aggregation (RA)}, to extend single-query to multi-query video retrieval. 
However, initial experiments with these approaches did not prove promising.
We believe this is because, while each new piece of information may provide complementary information, it may perform poorly as a query in isolation.
However, SA and RA work best with individual queries of high quality. 
We, therefore, adopt a different approach:
we aggregate pieces of information by concatenating them into one large query.
In this way, the initial query can be iteratively refined by adding more complementary information.
In detail, we concatenate multiple queries using a separation token \texttt{[SEP]}. 
This token is derived from BERT pretraining, in which it is used to separate sentences as follows: ``\texttt{[CLS]} sentence A \texttt{[SEP]} sentence B \texttt{[SEP]}'', suggesting its applicability for separating content in our application setting.

\subsection{Heuristic Question Generator}
\label{subsec:heuristic-question-generator}

We now turn the design of question generators used in our system, starting with a simple heuristic question generator.

\noindent
\textbf{Designing questions to improve retrieval performance}. 
The problem of asking useful questions about visual content has received attention in the field of Visual Question Generation. 
Mostafazadeh \text{\it{et al}}~\cite{mostafazadeh2016generating} proposed a framework that generates questions that can potentially engage a human in starting a conversation.
They observe that asking a question that can be answered simply by looking at the image would be of interest to the computer vision community, but those questions are neither natural nor engaging for a person to answer. 
Nevertheless, for the purpose of improving retrieval performance, we focus on concrete, discriminative questions (rather than abstract and engaging questions).
In particular, to develop a heuristic question generator, we target goal-driven questions~\cite{krishna2019information} that focus on extracting objects (``what is the person throwing?''), attributes (``what kind of shirt is the person wearing?''), color (``what color is the frisbee?''), material (``what material is the frisbee?''), etc. 

We are particularly interested in obtaining information that is highly discriminative and complementary to the original query.
Ask \& Confirm \cite{cai2021ask} argues that objects provide discriminative cues to distinguish a target from other candidates in the context of image retrieval.
Similarly, we observe the importance of the objects in the video retrieval task.
However, other cues such as color, human actions, and scenes are also important, and we, therefore, aim to construct a question generator that incorporates these cues. 

\noindent
\textbf{Heuristic Question Generator}. 
Our first approach to question generation employs hand-crafted heuristics.
Our strategy is to first obtain objects from the initial textual query (via pattern matching) and then ask questions about these objects. 
If an object is a living thing such as a person or animal, then we enquire about both the action and scene associated with it. 
If the object is a non-living thing such as a lamp or table, then we ask only the scene. 
If no object is found in the initial textual query, we ask the user to identify an object in the scene. 
The follow-up questions then proceed as above. 

Since living things are often associated with actions, 
if the object is a living thing, then we employ a template of the form ``who is doing what in what place''. 
If the object is a non-living thing, the template used is ``who is in what place''.
Once the answer is obtained, it is appended to the question, then both are appended to the current query for the next round of video retrieval.

One potential downside of this approach is the redundancy between questions and the query.
For example, suppose the initial textual query is ``a man is singing'', then when we ask ``what is the man doing?'', the answer may add no new information.
 In practice, this is typically not the case. 
We find that the information of the initial query is often restricted to addressing some temporal window within the video. 

However, the question provided to the user can often address a larger portion of the video.
For the example above, the man may walk through the street in the following frames after singing. 
The answer generator could encode this information to provide a complementary cue for retrieval.

\subsection{Parametric Question Generators}
\label{subsec:auto-question-generator}

The heuristic question generator captures intuition about what makes for useful questions but lacks flexibility.
We therefore explore question generation mechanisms that leverage a large language model that possesses the ability to ask flexible, open-ended questions.
The first, \textit{Auto-text}, asks questions based only on the text query.
The second, \textit{Auto-text-vid}, asks questions based on both the text query and the top-$K$ ranked video candidates for the current query (here $K$ is a hyperparameter of the system).

The foundation of both question generators is T-0++~\cite{sanh2022multitask}, a large language model that outperforms or matches GPT-3 \cite{brown2020language} on many NLP tasks.
Since this language model has been trained on various forms of question answering and shown to be very effective at zero-shot task generalization, we directly apply this model to generate questions without finetuning.

\noindent
\textbf{Auto-text}.
To generate questions from the query, we provide the following template to the language model: ``\textit{Suppose you are given the following video descriptions $Q_i$, What question would you ask to help you unique identify the video?}'' where $Q_0$ is the initial query and $Q_i$ is the concatenated query used to retrieve videos after the i$^{\mathrm{th}}$ round of interaction. 
The new query after each round incorporates the answer to the question through concatenation. Note that in terms of the Auto-text question generation approach, the question depends both on the original query and the new descriptions generated after each round of the interaction.

\noindent
\textbf{Auto-text-vid}. 
Differently from Auto-text, Auto-text-vid conditions its question generation on both the text query and the current top-$k$ ranked retrieved videos (selected via cosine similarity for the query corresponding to the current round of interaction).
This is because, intuitively, we expect the top-ranked videos selected at each round to share some common information with the target video. 
The use of the top-$k$ retrieved videos as conditioning information was also considered by~\cite{madasu2022learning}, who found it beneficial.

To proceed, we first generate captions for each of the top-ranked videos using the BLIP captioner~\cite{Li2022BLIPBL}. 
Then we use both these captions and the original text query to generate questions. 
The input sentence to the language model employs the following template: \textit{``Suppose you are given the following video descriptions: $C_i$. 
What question would you ask to help you unique identify the video described as follows: $Q_i$?''} where $C_i$ are the generated captions at the $i^{\mathrm{th}}$ round of interaction and $Q_i$ is the concatenated textual query used to retrieve videos after the $i^{\mathrm{th}}$ round of interaction. 
The questions generated are based on those common features, so the machine obtains more information about the target video through the interaction. 

\noindent \textbf{Remark.}
Since the top-ranked videos change at each round, there is some variation in the questions generated by~\textit{Auto-text-vid}.
This may provide complementary cues, but also carries the risk of additional noise (both through mistakes in the synthetic captions and through mismatches in content between the candidates and the target).

\subsection{Question Augmentation}
To further increase the information extracted from the user about the target video, we also introduce two further simple techniques to augment the questions: \textit{Ask Segment} (AS) and \textit{Ask Object} (AO). 

\noindent \textbf{Ask Segment} (AS) augmentation asks the user questions about \textit{each half of the video} independently. 
Videos evolve over time and different objects and actions may take place in different frames.
As a consequence, asking the same question about different halves of the video often yields very different answers.

\noindent \textbf{Ask Object} (AO) augmentation enquires about more objects in the scene. 
This approach is motivated by Ask \& Confirm~\cite{cai2021ask} in which Cai \text{\it{et al}} observe that objects are often highly discriminative.
Rather than searching for discriminative objects as candidates through a complex RL policy~\cite{cai2021ask}, we simply ask the user which objects are in the scene.

Note that these strategies can be combined to generate more discriminative questions for interactive video retrieval. 

\section{Experiment}

In this section, we first describe the datasets used in our experiments (\cref{subsec:datasets}).
We then discuss implementation details (\cref{subsec:implementation-details}) and evaluation metrics (\cref{subsec:evaluation}).

\subsection{Datasets}
\label{subsec:datasets}

\noindent
\textbf{MSR-VTT}~\cite{xu2016msr} is a video benchmark that is widely used for text-video retrieval. 
It contains 10,000 videos sourced from YouTube. 
Each video is annotated with 20 natural language descriptions.
Following previous work, we report retrieval results on the 1K-A test set. 

\noindent
\textbf{MSVD} \cite{chen-dolan-2011-collecting} is a dataset initially collected for translation and paraphrase evaluation. 
It contains 1,970 videos, spanning a duration from 1 second up to 62 seconds.
The train, validation, and test splits contain 1,200, 100, and 670 videos, respectively.
Each video contains approximately 40 associated sentences in English.

\noindent
\textbf{AVSD} (Audio-Visual Scene aware Dialog dataset)~\cite{alamri2019audio} is a collection of videos associated with ground truth dialog data.
Each video dialog consists of 10 rounds of human-generated questions and answers describing various details about the video content.
The AVSD dataset consists of 7985 training samples and 1863 validation samples. 
We follow prior work~\cite{madasu2022learning} and split the validation samples into a new validation split of 863 samples and 1000 test samples.
Note that our approach does not make use of the dialog data, since we simulate the user directly.
We use only the caption as the initial textual query to retrieve the video.

\subsection{Implementation Details}
\label{subsec:implementation-details}

\begin{figure*}[ht]
  \centering
   \includegraphics[width=\linewidth]{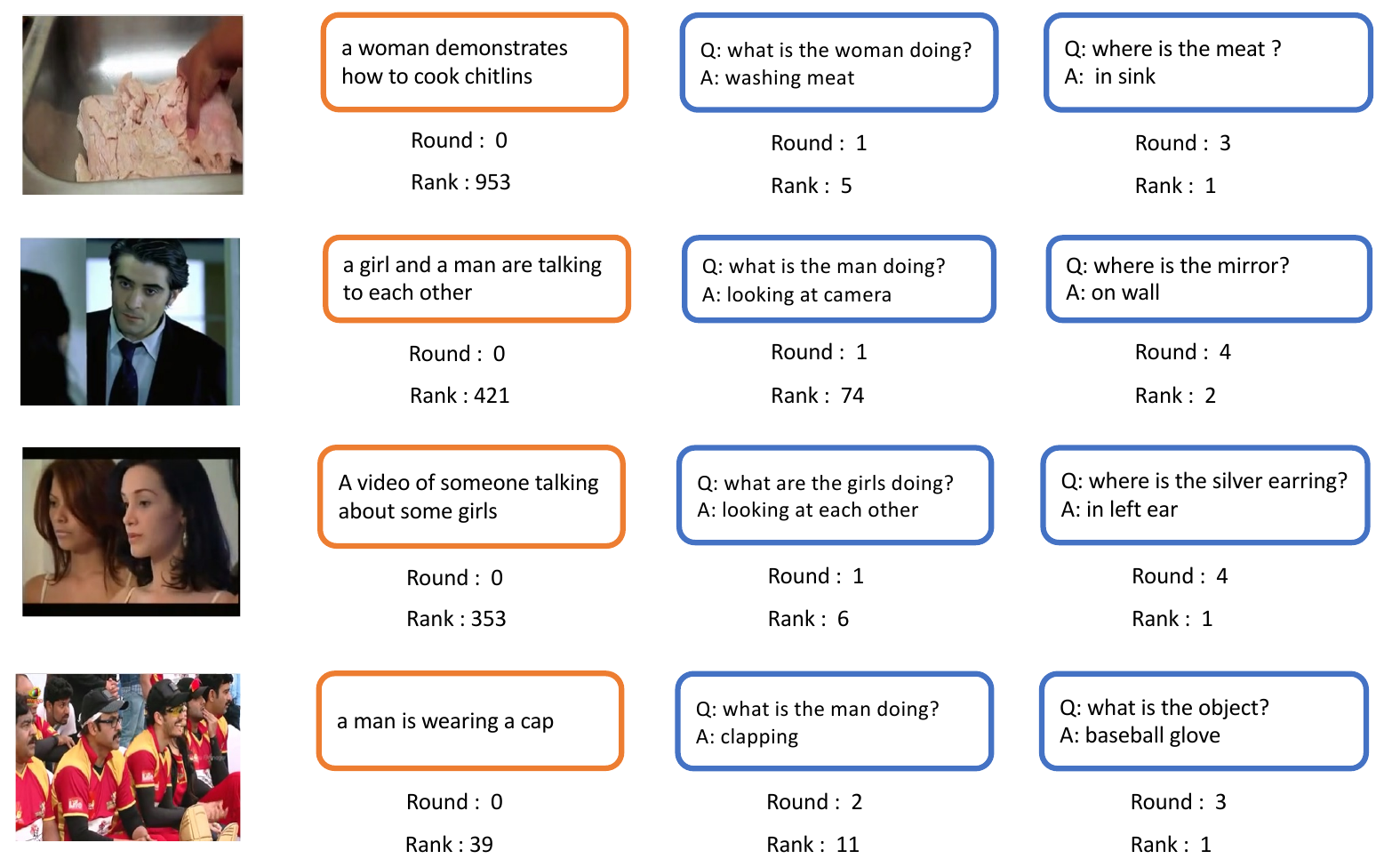}

   \caption{\textbf{Qualitative results of the dialogue generated by our Q \& A System on MSR-VTT.} 
   The questions are generated by the Heuristic Question Generator. 
   The initial queries of the target videos are shown in orange boxes. 
   The questions and answers generated through each interaction are shown in blue boxes. 
   Both the initial retrieval rank and the rank after the interaction are demonstrated.
   In each case, we observe a boost in performance.
   Note, however, that the answers are not always correct (e.g. in the fourth row ``A: baseball glove'' when the sport does not appear to be baseball).
   However, since the answer is ``nearly correct'', it still provides a useful cue to the retrieval model.}
   \label{fig:viz}
\end{figure*}

For the visual encoder, we employ a vision transformer with ViT-B/16 which contains 12 layers and 12 heads.
This model has a hidden dimensionality of 768, an MLP size of 3072, and an input patch size of $16 \times 16$. 
The text transformer used for text encoder is $\text{BERT}_{\text{BASE}}$ with 12 layers, 12 self-attention heads, and a hidden size of 768. 
The text transformer used for the image-grounded text encoder is the same as the text encoder except that it uses additional cross-attention layers to model vision-language interactions.
The architecture of the image-grounded text decoder is similar to the image-grounded text encoder but it replaces the bi-directional self-attention layers with causal self-attention layers.  
All videos are resized to $384 \times 384$ as input. 

In our interactive system, we adopt the BLIP retrieval model, the VQA model, and the captioner. 
The BLIP retrieval model is composed of an image encoder, a text encoder, and an image-grounded text encoder. 
The BLIP VQA model is composed of an image encoder, an image-grounded text encoder, and an image-grounded text decoder. 
The BLIP captioner contains an image encoder and an image-grounded text decoder.
To perform zero-shot transfer video retrieval, we uniformly sample 8 frames per video and concatenate the frame features into a single sequence. 
To generate questions, we adopt T0++ \cite{sanh2022multitask}, a powerful encoder-decoder language model trained on a collection of English-prompted datasets.
This model attains strong zero-shot task generalization on English natural language prompts (outperforming GPT-3 on many tasks while being 16x smaller). 

The Auto-text-vid question generation approach depends on the current top-k ranked retrieved videos. 
We experimented with different k (3, 5, 10, 15) and observed that when k = 5, we obtained the best performance. 
This may be because setting k to be too large produces too much noise in the generated captions while setting k to be too small does not provide an opportunity to extract all potentially discriminative information from related videos.

\noindent
\textbf{Baselines.} In addition to the BLIP baseline, we also experimented with the interactive approach in ~\cite{madasu2022learning} for fair comparisons. As public code is not available at the time of writing, we simulate their answer generation by first captioning the target video (by BLIP captioner) and then using the large language model (T0++) to generate the answer based on the caption and question. On the AVSD dataset, we also experimented with the approach in \cite{maeoki2020interactive, madasu2022learning} using 10 rounds of manually annotated human dialogue history for each video. However, simply concatenating all 10 rounds of ground truth dialog with the initial text query got a slightly worse performance than the baseline. Instead, we concatenated all the answers with the initial text query for retrieval.

\subsection{Evaluation}
\label{subsec:evaluation}
\noindent
\textbf{Metrics.} We adopt evaluation metrics that are widely used in the standard single-query retrieval setting and report recall@K (R@K, higher the better) and median rank (MdR, lower the better) in our experiments. 
R@K calculates the percentage of test data for which the ground-truth video is found in the retrieved K videos. 
MdR metric captures the median rank of the retrieved ground truth videos. 

\noindent
\textbf{Time cost.} We next assess the time cost of producing answers during the interaction.
For this, we sample 50 videos from the MSR-VTT dataset. We aim to compare our answer generation approach using the VideoQA model with the approach in ViReD ~\cite{madasu2022learning} (We call this answer generation approach \textit{CAP+LM}). 
In addition, we also conduct a user study on answer generation.
We invite 4 users to answer the auto-generated question (Auto-text) on the target videos and record the time they spent on each question. (Note that each user went through all 50 videos before the interaction.)

\section{Results}

\begin{table}[ht]
\centering
\setlength{\tabcolsep}{0.4em} 
\renewcommand{\arraystretch}{1.2}
\scalebox{0.85}{
\begin{tabular}{ c|c|c c c c} 
 \hline
 Method & ITA & R1 $\uparrow$ & R5 $\uparrow$ & R10 $\uparrow$ & MdR $\downarrow$ \\
 \hline
 FiT \cite{bain2021frozen} & $\times$ & 31.9 & 61.1 & 72.6 & 3 \\
 CLIP4Clip \cite{luo2021clip4clip} & $\times$ & 41.5 & 69.4 & 79.3 & 2 \\
 CenterCLIP \cite{zhao2022centerclip} & $\times$ & 48.4 & 73.8 & 82.0 & 2 \\
 X-CLIP \cite{ma2022x} & $\times$ & 49.3 & 75.8 & 84.8 & 2 \\
 \hline
 BLIP \cite{Li2022BLIPBL} & $\times$ & 43.3 & 65.6 & 74.7 & 2 \\
 BLIP + ViReD \cite{madasu2022learning} & $\surd$ & 47.0 & 72.3 & 81.4 & 2 \\
\hline
 BLIP + Auto-text w/o AO & $\surd$ & 47.6 & 72.8 & 81.8 & 2 \\
 BLIP + Auto-text-vid w/o AO & $\surd$ & 49.0 & 72.9 & 80.0 & 2 \\
 BLIP + Auto-text & $\surd$ & 62.2 & 84.5 & 90.7 & 1 \\
 BLIP + Auto-text-vid & $\surd$ & 57.6 & 82.2 & 88.7 & 1 \\
 BLIP + Heuristic  & $\surd$ & \bf{67.9} & \bf{89.5} & \bf{94.9} & \bf{1} \\
 \hline
\end{tabular}}
\vspace{2mm}
\caption{Comparison to state-of-the-art methods for text-to-video retrieval on the 1k test split of MSRVTT dataset. ITA indicates whether it is an interactive retrieval system. Recall metrics are reported in percent.
\label{tab:msr-vtt}}
\end{table}

\begin{table}[htb]
\centering
\setlength{\tabcolsep}{0.4em} 
\renewcommand{\arraystretch}{1.2}
\scalebox{0.86}{
\begin{tabular}{ c|c|c c c c} 
 \hline
 Method & ITA & R@1 $\uparrow$ & R@5 $\uparrow$ & R@10 $\uparrow$ & MdR $\downarrow$ \\
 \hline
 FiT \cite{bain2021frozen} & $\times$ & 33.7 & 64.7 & 76.3 & 3 \\
 CLIP4Clip \cite{luo2021clip4clip} & $\times$ & 46.2 & 76.1 & 84.6 & 2 \\
 X-CLIP \cite{ma2022x} & $\times$ & 50.4 & 80.6 & - & - \\
 CenterCLIP \cite{zhao2022centerclip} & $\times$ & 50.6 & 80.3 & 88.4 & 1 \\
 \hline
 BLIP \cite{Li2022BLIPBL} & $\times$ & 44.2 & 73.7 & 80.9 & 2 \\
 BLIP + ViReD \cite{madasu2022learning} & $\surd$ & 51.0 & 78.1 & 87.2 & 1 \\
 \hline
 $\text{BLIP}$ + Auto-text & $\surd$  & 68.7 & 91.2 & 95.8 & 1 \\
 $\text{BLIP}$ + Auto-text-vid & $\surd$  & 66.1 & 88.8 & 93.7 & 1 \\
 $\text{BLIP}$ + Heuristic & $\surd$ & \bf{74.2} & \bf{93.4} & \bf{97.9} & \bf{1} \\
 \hline
\end{tabular}}
\vspace{2mm}
\caption{Comparison to state-of-the-art methods for text-to-video retrieval on the test split of the MSVD dataset. ITA indicates whether it is an interactive retrieval system.
\label{tab:msvd}}

\end{table}

\begin{table}[htb]
\centering
\setlength{\tabcolsep}{0.4em} 
\renewcommand{\arraystretch}{1.2}
\scalebox{0.81}{
\begin{tabular}{ c|c|c c c c } 
 \hline
 Method & ITA & R@1 $\uparrow$ & R@5 $\uparrow$ & R@10 $\uparrow$ & MdR $\downarrow$ \\
 \hline
 FiT \cite{bain2021frozen} & $\times$ &  5.6 & 18.4 & 27.5 & 25 \\
 LSTM \cite{maeoki2020interactive} &  $\surd$ & 4.2 & 13.5 & 22.1 & - \\
 FiT + Human dialog \cite{madasu2022learning} & $\surd$ & 10.8 & 28.9 & 40.0 & 18 \\
 ViReD \cite{madasu2022learning} & $\surd$ & 24.9 & 49.0 & 60.8 & 6 \\
 \hline
 BLIP \cite{Li2022BLIPBL} & $\times$ & 27.3 & 50.0 & 61.2 & 5.5 \\
 BLIP + Human dialog \cite{madasu2022learning} & $\surd$ & 27.4 & 50.4 & 62.7 & 5 \\
 BLIP + ViReD \cite{madasu2022learning} & $\surd$ & 28.4 & 54.3 & 66.8 & 4 \\
 \hline
 BLIP + Auto-text  & $\surd$ & 36.0 & 61.6 & 71.6 & 3 \\
 BLIP + Auto-text-vid  & $\surd$ & 36.5 & 61.0 & 71.1 & 3 \\
 BLIP + Heuristic & $\surd$ & \bf{39.5} & \bf{69.1} & \bf{77.8} & \bf{2} \\
 \hline
\end{tabular}}
\vspace{2mm}
\caption{Comparison to state-of-the-art methods for text-to-video retrieval on the test split of AVSD dataset. ITA indicates whether it is an interactive retrieval system.
\label{tab:avsd}}

\end{table}

\begin{table}[htb]
\centering
\setlength{\tabcolsep}{0.4em} 
\renewcommand{\arraystretch}{1.2}
\scalebox{0.9}{
\begin{tabular}{c|c c c } 
 \hline
 & VideoQA & CAP+LM & User \\
 \hline
 Avg time (s) & 0.1 & 8.0 & 5.7\\
 \hline
\end{tabular}}
\vspace{2mm}
\caption{Comparison of the average time cost of different answer and question generation approaches. We show the results of the answer generation approach of VQA (ours) and CAP+LM (simulation of ViReD \cite{madasu2022learning} answer generator) and the user study. 
\label{tab:time}}

\end{table}

\begin{figure}[ht]
  \centering
   \includegraphics[width=\linewidth]{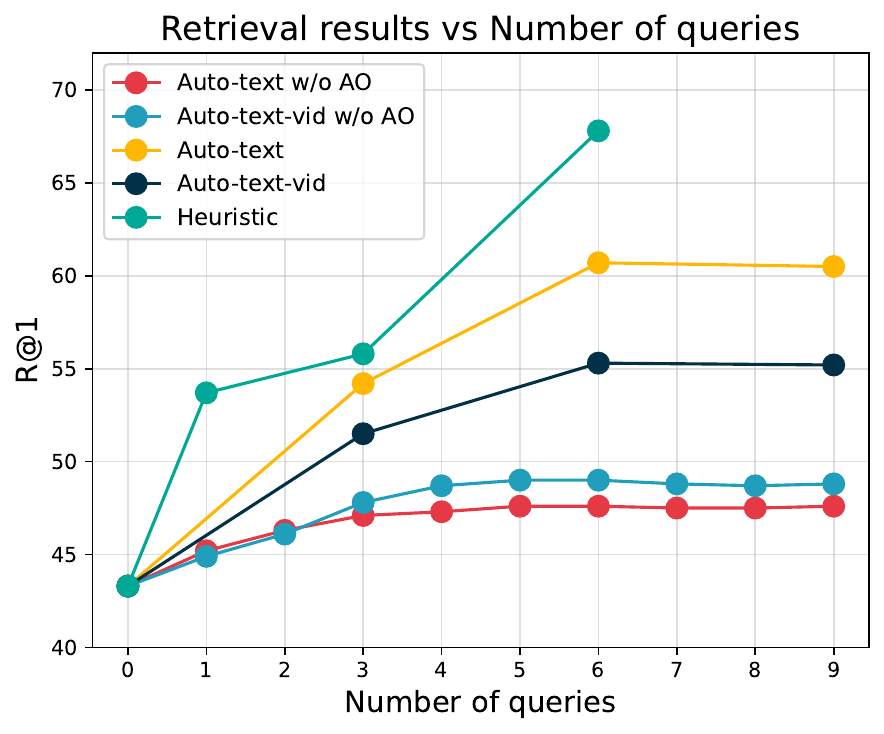}
   \caption{\textbf{Retrieval result vs Number of queries on MSRVTT.} Performance of all different approaches improves as the number of rounds increases. Heuristic question generation approach achieves the most significant improvement.}
   \label{fig:numq}
\end{figure}

In this section, we discuss experimental results.
On MSR-VTT (\cref{tab:msr-vtt}), without AO (asking objects), we observe that both the Auto-text and Auto-text-vid question generation approaches outperform the BLIP baseline.
We observe that Auto-text-vid, which makes use of the top-ranked retrieved videos at each round, performs better than Auto-text.
With AO, there is a significant performance boost for both approaches.
For instance, we observe that the Auto-text \QA System outperforms the baseline by +18.9\% in recall@1.
Interestingly, we observe that Auto-text outperforms Auto-text-vid when used in combination with AO.
This suggests that simply asking the user directly for the objects that they are interested in (rather than inferring them from closely ranked videos) offers higher signals.

We also observe that the Heuristic question generator yields dramatic gains:
the recall@1 is improved by 24.6\% (from 43.3\% to 67.9\%), 30\% (from 44.2\% to 74.2\%) and 12.2\% (from 27.3\% to 39.5\%) on MSR-VTT, MSVD and AVSD respectively. 
Using Auto-text question generator, the BLIP baseline recall@1 is improved by 18.9\% (from 43.3\% to 62.2\%), 24.5\% (from 44.2\% to 68.7\%), and 8.7\% (from 27.3\% to 36.0\%) on MSR-VTT, MSVD, and AVSD respectively. 
This indicates that generated questions elicit useful information from the (simulated) user that substantially improves retrieval performance.
Compared to prior interactive approaches (ViReD, Human dialog) on AVSD, our approach does not use dialog training data for finetuning but nevertheless outperforms existing work.
Interestingly, the heuristic approach performs best over all three different datasets.

As seen in \cref{fig:viz}, a few rounds of interaction are sufficient to locate the large video. We demonstrate the cases where the initial query is incomplete, hindering performance. However, using the question-answering system, the search results quickly improve and we obtain significantly better retrieval performance. 
The generated questions are of high quality, aiming to confirm discriminative information from the video. 
In addition, the answers generated by VideoQA model are reasonable in the majority of cases (though not always perfect).
They successfully capture different aspects of the information in the video (such as action, location, and object). Another useful property of the generated answers is that they are short phrases, simulating concise user answers. 

\noindent
\textbf{Number of queries.} 
Additional rounds of interaction lead to additional new queries.
We observe in \cref{fig:numq} that the retrieval performance increases as the number of queries increases. 
This effect is clearest for Heuristic question generation approach. 
The performance of both Auto-text and Auto-text-vid increases faster than their variants without AO. 
This further confirms that asking users about objects in the videos is an effective approach to improve retrieval performance at each round of the interaction. 
We observe that without AO, recall@1 of Auto-text-vid increases faster than that of Auto-text while the opposite result is observed with AO. This agrees with our argument ``simply asking the user directly for the objects that they are interested in (rather than inferring them from closely ranked videos) offers higher signal''. For all the approaches, Performance stabilizes as the number of queries grows beyond 6. Note that the maximum number of queries for heuristic question generation is fixed to 6 and using 6 rounds works best.

\noindent
\textbf{Computational cost.} As shown in \cref{tab:time}, Our approach using the VideoQA model is significantly more efficient than CAP+LM. This is as expected because two steps (captioning and question-answering) are involved in CAP+LM approach while VideoQA directly incorporates the video and question information to generate answers. 
In our naive implementation, CAP+LM is even slower than the user response (though this could be optimised).

\section{Discussion}

\noindent
\textbf{Limitations.}  
(1) T0++ \cite{sanh2021multitask} is good at question-answering tasks but might not be good at generating questions. Therefore fine-tuning the model for the question generation task could be helpful.
(2) We adopt the BLIP model for both retrieval and VideoQA. 
In principle, there is the potential for ``leakage'' in which the VideoQA model encodes information that is specific to BLIP, but irrelevant to the video.
In practice, different parameter checkpoints are used for different VideoQA and retrieval components, so we believe that such leakage is unlikely to occur in practice.

\noindent \textbf{Broader Impact.}
Improved video retrieval has a wide range of societally useful applications, spanning domains such as education, entertainment, and security.
However, as with many technologies, it is subject to dual use. 
For instance, it could be used to facilitate unlawful surveillance.
\section{Conclusion}

In this work, we proposed several simple baselines for interactive video retrieval based on the generation of questions and answers.
Through experiments on standard benchmarks, we demonstrated the effectiveness of generated questions in combination with a VideoQA model to simulate user responses through the interaction process.
In future work, we hope to explore the use of question-answering systems to simulate users in other retrieval tasks.

\noindent \textbf{Acknowledgements.}
\noindent \noindent SA would like to acknowledge Z. Novak and N. Novak in enabling his contribution, and support from an Isaac Newton grant and an EPSRC HPC grant. KL would like to thank Bill Byrne, Olga Russakovsky, Yu Wu, Vishaal Udandarao, Allison Chen, Xindi Wu, and Zhiwei Deng for helpful discussions and feedback.

{\small
\bibliographystyle{ieee_fullname}
\bibliography{reference}
}

\newpage
\appendix

In the supplementary material, we report results on additional datasets (\cref{sec:more_results}), analyze failure cases of our interactive system (\cref{sec:failure_cases}), provide additional results on the number of queries (\cref{sec:num-queries}), and evaluate our system with user study (\cref{sec:user-study}).

\section{Results on additional datasets}
\label{sec:more_results}
\begin{table}[ht]
\begin{center}
\setlength{\tabcolsep}{0.4em} 
\renewcommand{\arraystretch}{1.2}
\scalebox{0.9}{
\begin{tabular}{ c|c|c c c c} 
 \hline
 Method & ITA & R1 $\uparrow$ & R5 $\uparrow$ & R10 $\uparrow$ & MdR $\downarrow$ \\
 \hline
 CLIP4Clip \cite{luo2021clip4clip} & $\times$ & 21.6 & 41.8 & 49.8 & 11 \\
 BLIP  \cite{Li2022BLIPBL} & $\times$ & 21.3 & 38.2 & 45.3 & 16 \\
\hline
 BLIP + Auto-text & $\surd$ & 41.2 & 63.8 & 73.1 & 2 \\
 BLIP + Auto-text-vid & $\surd$ & 39.1 & 61.7 & 70.4 & 3 \\
 BLIP + Heuristic  & $\surd$ & \bf{55.7} & \bf{78.0} & \bf{85.9} & \bf{1} \\
 \hline
\end{tabular}}
\caption{Text-to-video retrieval results on LSMDC dataset. ITA
indicates whether it is an interactive retrieval system.
\label{tab:lsmdc}}
\end{center}
\end{table}
\vspace{-3mm}

\begin{table}[ht]
\begin{center}
\setlength{\tabcolsep}{0.4em} 
\renewcommand{\arraystretch}{1.2}
\scalebox{0.9}{
\begin{tabular}{ c|c|c c c c} 
 \hline
 Method & ITA & R1 $\uparrow$ & R5 $\uparrow$ & R10 $\uparrow$ & MdR $\downarrow$ \\
 \hline
 CLIP4Clip \cite{luo2021clip4clip}  & $\times$ & 43.4 & 70.2 & 80.6 & 2 \\
 BLIP  \cite{Li2022BLIPBL} & $\times$ & 41.5 & 67.1 & 75.3 & 2 \\
\hline
 BLIP + Auto-text & $\surd$ & 50.6 & 77.5 & 85.5 & 1 \\
 BLIP + Auto-text-vid & $\surd$ & 45.0 & 70.8 & 79.2 & 2 \\
 BLIP + Heuristic  & $\surd$ & \bf{61.6} & \bf{86.9} & \bf{91.5} & \bf{1} \\
 \hline
\end{tabular}}
\caption{Text-to-video retrieval results on DiDeMo dataset. ITA
indicates whether it is an interactive retrieval system.
\label{tab:didemo}}
\end{center}
\end{table}

\vspace{-3mm}
\begin{table}[ht]
\begin{center}
\setlength{\tabcolsep}{0.4em} 
\renewcommand{\arraystretch}{1.2}
\scalebox{0.9}{
\begin{tabular}{ c|c|c c c c} 
 \hline
 Method & ITA & R1 $\uparrow$ & R5 $\uparrow$ & R10 $\uparrow$ & MdR $\downarrow$ \\
 \hline
 CLIP4Clip \cite{luo2021clip4clip} & $\times$ & 40.5 & 72.4 & - & 2 \\
 BLIP \cite{Li2022BLIPBL} & $\times$ & 35.3 & 60.5 & 72.4 & 3 \\
\hline
 BLIP + Auto-text & $\surd$ & 40.5 & 65.0 & 75.0 & 2 \\
 BLIP + Auto-text-vid & $\surd$ & 35.4 & 59.3 & 69.9 & 3 \\
 BLIP + Heuristic  & $\surd$ & \bf{44.8} & \bf{70.4} & \bf{79.9} & \bf{2} \\
 \hline
\end{tabular}}
\caption{Text-to-video retrieval results on ActivityNet. ITA
indicates whether it is an interactive retrieval system.
\label{tab:activity}}
\end{center}
\end{table}

We further evaluated our method using three additional widely recognized video retrieval datasets: LSMDC \cite{rohrbach2015dataset}, DiDeMo \cite{anne2017localizing}, and ActivityNet \cite{caba2015activitynet}. For DiDeMo and ActivityNet, we set up the baseline using the same approach as in CLIP4Clip \cite{luo2021clip4clip}. In our approach, we used the first sentence of the entire caption as the initial query and iteratively refined it through interaction. The results clearly demonstrate the effectiveness of our approach, as it consistently outperformed the baseline method across all three datasets. Particularly noteworthy is the fact that although the Heuristic question generator is manually designed, it exhibits remarkable robustness.

We noticed that the improvement on ActivityNet is not as pronounced as in the other datasets, which can be attributed to the limitations of BLIP \cite{Li2022BLIPBL}. BLIP's lack of pre-training on video datasets, as well as its inability to utilize temporal information, pose significant challenges in long-form video understanding within the zero-shot retrieval setting. Nevertheless, our approach still demonstrates a substantial improvement over the baseline, showcasing its effectiveness.

\section{Analyze failure cases}
\label{sec:failure_cases}
\begin{figure*}[ht]
  \centering
   \includegraphics[width=\linewidth]{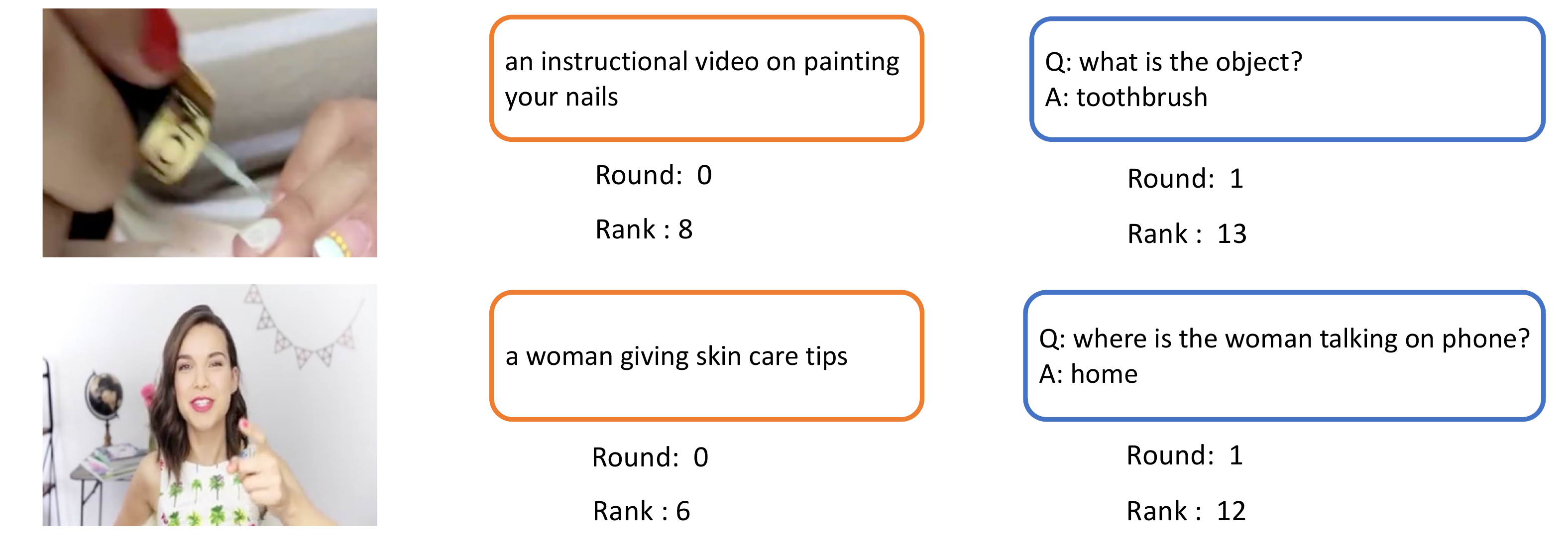}
   \caption{\textbf{Qualitative results of the failure cases using our Q \& A System on MSR-VTT.} 
   The questions are generated by the Heuristic Question Generator. 
   The initial queries of the target videos are shown in orange boxes. 
   The questions and answers generated through each interaction are shown in blue boxes. 
   Both the initial retrieval rank and the rank after the interaction are demonstrated.}
   \label{fig:fail}
\end{figure*}

It has been observed that the failures in the performance of VideoQA models can be attributed to the limitations of both the question and answer generators. To illustrate, in the first example depicted in \cref{fig:fail}, the VideoQA model fails to identify the object in the video (the object is not a toothbrush). In the second example, the question generated is misleading as the woman in the video is not talking on the phone. Furthermore, the generated questions are at times not meaningful, as the answer to the question is already present in the initial query. In such cases, the performance of our system remains unchanged or may even slightly deteriorate.

\section{The influence of the number of queries (recall@5)}
\label{sec:num-queries}

\begin{figure}[ht]
  \centering
   \includegraphics[width=\linewidth]{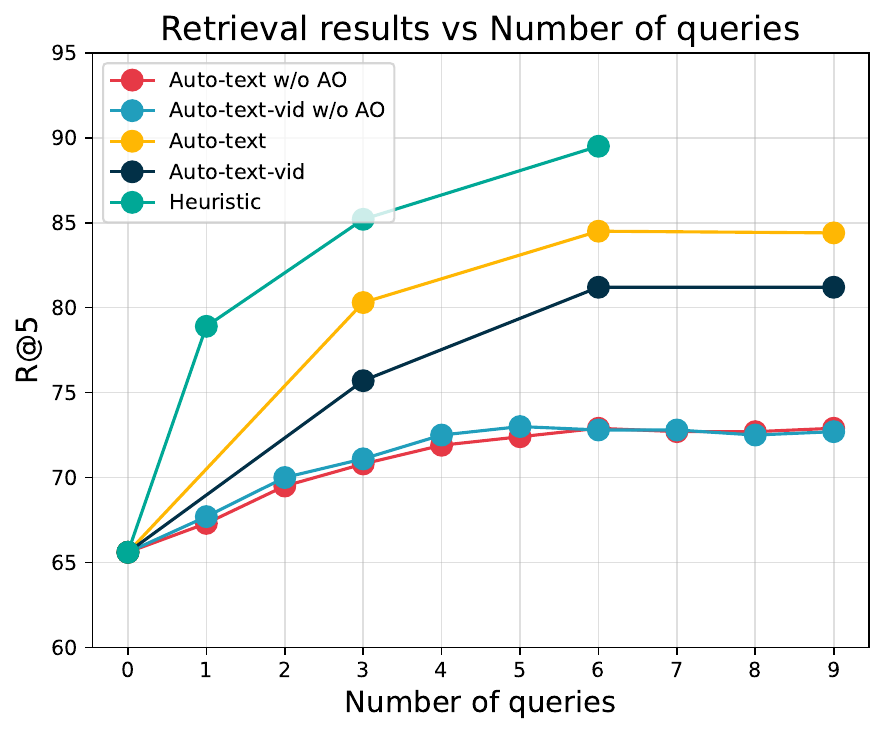}
   \caption{\textbf{Recall@5 vs number of queries on MSRVTT.}
   The performance of all approaches improves as the number of queries increases and then stabilizes as the number of queries grows beyond 6. The maximum number of queries for heuristic question generation is fixed to be 6.}
   \label{fig:numq}
\end{figure}
To supplement the main paper, we provide additional results illustrating how the number of queries influences retrieval performance under a recall@5 metric (~\cref{fig:numq}).
We observe a similar trend to the recall@1 results reported in the main paper.

\section{User Study}
\label{sec:user-study}

\begin{table}
\begin{subtable}[ht]{0.5\textwidth}

\centering
\renewcommand{\arraystretch}{1.2}
\begin{tabular}{ c|c c c c} 
 \hline
 Method & VideoQA & CAP+LM & User\\
 \hline
 Heuristic & 66.0 & 60.0 & 69.0 \\
 Auto-text & 66.0 & 62.0 & 68.0 \\
 Auto-text-vid & 62.0 & 56.0 & 66.0 \\
 \hline
\end{tabular}
\vspace{2mm}
\caption{Recall@1
\label{tab:1}}
\end{subtable}
\vspace{2mm}

\begin{subtable}[ht]{0.5\textwidth}
\centering
\renewcommand{\arraystretch}{1.2}

\begin{tabular}{ c|c c c c} 
 \hline
 Method & VideoQA & CAP+LM & User\\
 \hline
 Heuristic & 90.0 & 84.0 & 91.5\\
 Auto-text & 80.0 & 78.0 & 84.5 \\
 Auto-text-vid & 82.0 & 76.0 & 84.0 \\
 \hline
\end{tabular}
\vspace{2mm}
\caption{Recall@5
\label{tab:2}}
\end{subtable}
\vspace{2mm}

\begin{subtable}[ht]{0.5\textwidth}
\centering
\renewcommand{\arraystretch}{1.2}

\begin{tabular}{ c|c c c c} 
 \hline
 Method & VideoQA & CAP+LM & User\\
 \hline
 Heuristic & 96.0 & 90.0 & 95.0 \\
 Auto-text & 88.0 & 86.0 & 90.0 \\
 Auto-text-vid & 88.0 & 88.0 & 91.0 \\
 \hline
\end{tabular}
\vspace{2mm}
\caption{Recall@10
\label{tab:3}}
\end{subtable}
\caption{
\textbf{Comparing VideoQA answers to human answers on a selection of 50 randomly sampled videos in MSR-VTT}. 
Recall metrics are reported as percent.
The retrieval results are averaged across 4 different users.
We observe that VideoQA performs similarly to human answers.}
\label{tab:user_study}

\end{table}
We conducted a user study to confirm that our VideoQA model provides a reasonable simulation of human responses.
Four volunteers (graduate students) were invited to participate in our interactive video retrieval experiments. 
For this study, we randomly selected 50 videos from the MSR-VTT dataset. 
Each participant watched the videos and provided short answers to the generated questions.
All three question generation approaches (Heuristic, Auto-text, and Auto-text-vid) were considered. 
The final retrieval results are averaged across 4 different users.

As seen in \cref{tab:user_study}, the VideoQA model obtains similar performance to the user, indicating that the VideoQA model provides a reasonable approximation. 
In addition, the CAP+LM model consistently performs worse than the VideoQA model. 
This result is expected since when a system generates a caption, it does not know what the question will be. 
As such, it is likely to miss relevant details.
However, the VideoQA model (which incorporates the information from both videos and questions) addresses this issue.

\end{document}